\documentclass[letterpaper]{article} % DO NOT CHANGE THIS
\usepackage{aaai2027}  % DO NOT CHANGE THIS
\usepackage{times}  % DO NOT CHANGE THIS
\usepackage{helvet}  % DO NOT CHANGE THIS
\usepackage{courier}  % DO NOT CHANGE THIS
\usepackage[hyphens]{url}  % DO NOT CHANGE THIS
\usepackage{graphicx} % DO NOT CHANGE THIS
\usepackage{natbib}  % DO NOT CHANGE THIS + REMOVE UNUSED CITATIONS LATER
\usepackage{caption} % DO NOT CHANGE THIS
\usepackage{booktabs}
\usepackage{multirow}
\usepackage{amssymb}
\usepackage{amsmath}
\usepackage{booktabs}
\usepackage{multicol}
\usepackage{multirow}
\usepackage{algorithm}
\usepackage{algorithmic}
\usepackage{pifont}
\newcommand{\cmark}{\ding{51}} % ✔
\title{CoMem: Collective-Individual Memory Synergy for Evolutionary Multi-Agent Systems}
\author{
    Chengxin Yu\textsuperscript{1},
    Zhaoxin Fan\textsuperscript{1}\textsuperscript,
    Faguo Wu\textsuperscript{1},
    Hongwei Zheng\textsuperscript{2},\\
    Yun Zhou\textsuperscript{3},
    Zhiyu Li\textsuperscript{4}
}
\affiliations{
    \textsuperscript{1} Beijing Advanced Innovation Center for Future Blockchain and Privacy Computing, School of Artificial Intelligence, Beihang University\\
    \textsuperscript{2} Beijing academy of blockchain and edge computing\\
    \textsuperscript{3} National University of Defense Technology\\
    \textsuperscript{4} MemTensor (Shanghai) Technology Co., Ltd.\\
}
\nocopyright
\begin{document}

\maketitle

\begin{abstract}
Designing effective memory mechanisms is crucial for advancing LLM-driven Multi-Agent Systems (MAS), helping agents learn together and perform better over time. While recent work has led to strong cooperation skills, most methods still use flat, unstructured memories, which easily get filled with noise and erase differences between agents. To address this, we introduce the concept of collective-individual memory synergy and propose CoMem, an architecture that unifies both private experience and shared knowledge for multi-agent learning. CoMem features:
(i) Private Experience Sedimentation, which lets each agent keep and update its own useful memories over time;
(ii) Collective Wisdom Curation, which carefully selects only widely proven ideas to be shared among agents;
(iii) Parallel Dual-Stream Retrieval, which allows agents to draw both from their own memory and the group's wisdom, using clustering to ensure diversity.
Experiments on ALFWorld and PDDL benchmarks show that CoMem achieves strong overall performance and robustly avoids memory pollution.
\end{abstract}

% %%%%%%%%%%%%%%%%%%%%%%%%%%%%%%%%%%%%%%%%%%%%%%%%%%%%%
% SECTION 1: INTRODUCTION
% %%%%%%%%%%%%%%%%%%%%%%%%%%%%%%%%%%%%%%%%%%%%%%%%%%%%%
\section{Introduction}

Multi-Agent Systems (MAS) have emerged as a key paradigm for tackling complex, real-world problems that require coordination, cooperation, and collective problem-solving\cite{wooldridge2009}. Recent advances in Large Language Models (LLMs)\cite{react2023, selfrefine2023, xi2025} have greatly enhanced these systems, enabling autonomous agents to reason, plan, and interact more effectively in diverse and dynamic environments\cite{autogen2024, dylan2024, macnet2025, guo2024, chen2024, han2024, du2024debate, moa2024}. As MAS are deployed in increasingly challenging scenarios, it becomes essential for agents not only to act collaboratively, but also to learn from past experiences—highlighting the critical role of effective memory mechanisms in supporting continual adaptation and long-term group intelligence, which we call group memory\cite{amem2025, mem02025, awm2025}.

\begin{figure}[t]
\centering
\includegraphics[width=1\columnwidth]{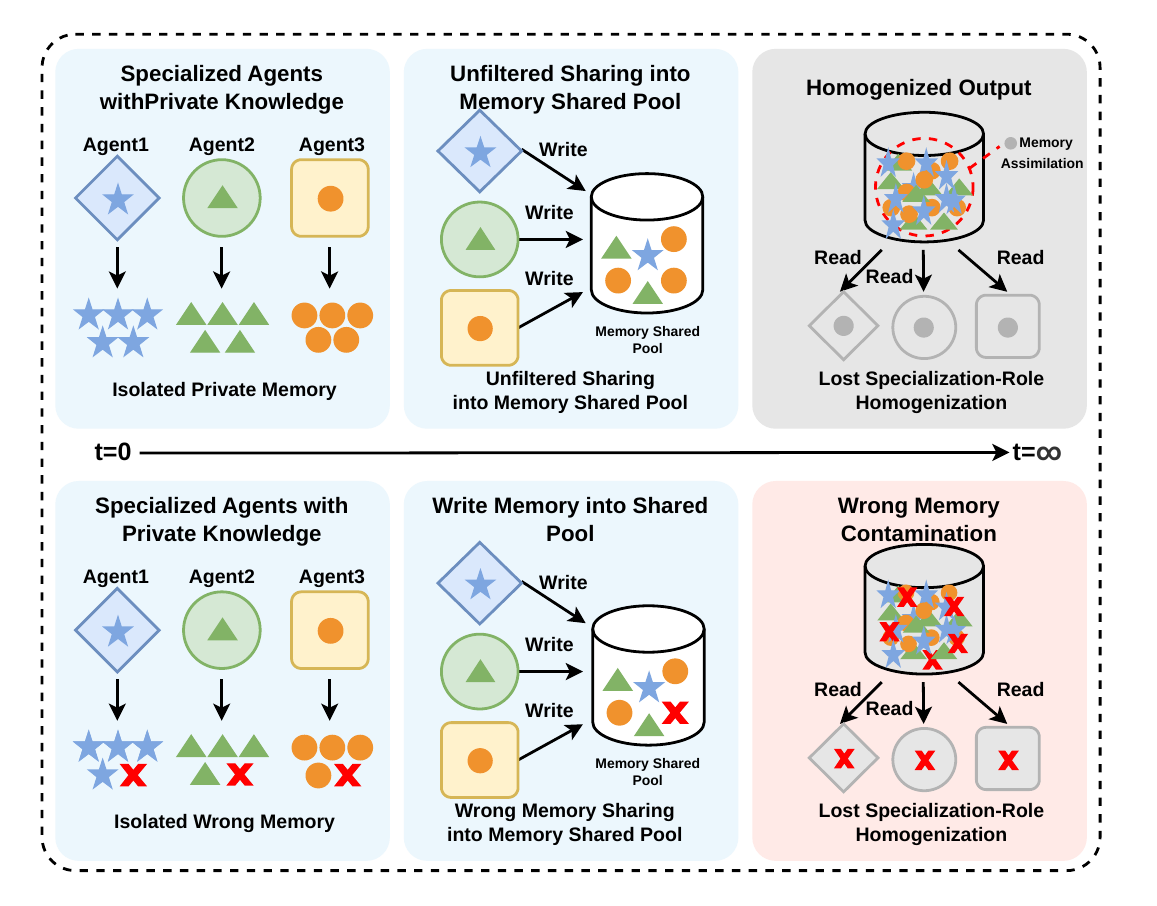}
\caption{Illustration of memory pollution in MAS. We conceptualize the problem as comprising two intertwined degradation processes: (a) noise accumulation, where unverified trial-and-error behaviors and low-quality decisions overwhelm the shared repository, and (b) behavioral homogenization, where heterogeneous agents lose their role-specific characteristics through blind cross-writing. These two processes mutually reinforce each other, ultimately driving the collective toward mediocrity.}
\label{fig:architecture}
\end{figure}

This importance of group memory in MAS is evident in everyday life. For instance, people who come from the same cultural background—having heard the same childhood stories or traditions—often find it easier to understand each other and work together. Such collective memories create a sense of belonging and common ground, fostering greater trust and cohesion when tackling problems as a group\cite{khushiyant2025, woolley2010}.

Recent work has begun to explore memory mechanisms for MAS, such as introducing shared memory spaces that enable agents to store and retrieve strategic experiences across tasks~\cite{gmemory2025}. However, these approaches typically employ simple, flat memory structures without explicitly modeling the concept of group memory or considering the synergy between collective and individual memories outlined above\cite{zhang2025, liu2025, du2026}. As a result, such designs often overlook the need for coordination and quality control in shared memory, making them susceptible to memory pollution—a phenomenon illustrated in Figure~\ref{fig:architecture}. This “memory pollution” mainly appears in two forms: (i) noise accumulation, where unverified or low-quality agent behaviors flood the shared memory and reduce its reliability; and (ii) behavioral homogenization, where agents lose their individual specializations as their experiences are repeatedly averaged out\cite{wang2026, bras2026}. These issues reinforce each other and often lead to worse performance in flat memory designs, rather than the intended improvement.

To tackle the issue, we argue that sustainable collective intelligence in multi-agent systems requires more than shared memory: it demands a careful balance between robust individual learning and meaningful group knowledge to construct group memory\cite{du2026, rezazadeh2025, amem2025}. Mirroring the dynamics of human societies, we hypothesize that effective population-level memory emerges only when individual agents are able to learn, test, and refine experiences in isolation, while genuine consensus is distilled through selective sharing and rigorous screening. In this way, collective progress is grounded in the diversity and integrity of private exploration, with only the most reliable insights crystallizing at the group level to inform future decision-making.

Inspired by the knowledge evolution process observed in human communities—“individual proposal, collective screening, consensus crystallization”~\cite{bruggeman2023collective}—we model group memory as a tightly coupled ecosystem with two distinct components: a private layer for interference-free accumulation of agent-specific experience, and a curated collective pool for distilled group wisdom. These layers interact through a structured promotion pathway: private memory retains only frequently validated knowledge through usage-aware rolling pruning, while collective memory admits new insights only after repeated, cross-agent verification. This bidirectional flow forms a virtuous cycle, where trustworthy private discoveries are promoted upward and refined collective knowledge guides subsequent individual behavior.

To operationalize this principle, we introduce {CoMem}, a novel memory architecture designed to support continual agent learning while suppressing memory pollution. CoMem centers on three synergistic mechanisms: (i) Private Experience Sedimentation with rolling pruning, allowing each agent to preserve and refine a pure history of its own validated experiences; (ii) Collective Wisdom Curation, employing a stringent promotion process so that only knowledge proven robust across tasks enters the shared pool; and (iii) Parallel Dual-Stream Retrieval, enabling each agent to blend private and collective memories for decision-making, while maintaining diversity through clustering-based filtering\cite{lu2026}. Together, these components establish a structured and dynamic memory ecosystem that enables both individual exploration and lasting group intelligence. Experiments on ALFWorld and PDDL benchmarks demonstrate the effectiveness of CoMem, achieving consistent gains over strong memory-augmented MAS baselines. 

Our main contributions are as follows:
\begin{itemize}
\item We propose CoMem, a novel memory architecture designed to mitigate memory pollution in multi-agent systems with flat group memory.
\item CoMem introduces a collective-individual memory synergy with key components: private experience sedimentation, collective wisdom curation, and parallel dual-stream retrieval, structurally balancing individual and group learning.
\item Extensive experiments on ALFWorld and PDDL benchmarks demonstrate that CoMem consistently improves over strong baselines while effectively mitigating memory pollution.
\end{itemize}
% %%%%%%%%%%%%%%%%%%%%%%%%%%%%%%%%%%%%%%%%%%%%%%%%%%%%%
% SECTION 2: RELATED WORK
% %%%%%%%%%%%%%%%%%%%%%%%%%%%%%%%%%%%%%%%%%%%%%%%%%%%%%
\section{Related Work}
\subsection{LLM-based Multi-Agent Systems}
Recent advances have expanded LLMs from single-turn language generators into autonomous agents capable of independent reasoning, planning, and tool utilization~\cite{brooks1991, react2023, xi2025}. Building on this individual empowerment, MAS orchestrate multiple LLM agents to tackle intricate tasks via collaboration\cite{du2024debate, moa2024}. Frameworks like AutoGen~\cite{autogen2024} and CAMEL~\cite{camel2023} introduce customizable agent communication topologies, while ChatDev~\cite{chatdev2024} and MetaGPT~\cite{metagpt2024} simulate software development workflows using role-specific agents. To optimize collaborative dynamics, DyLAN~\cite{dylan2024} dynamically selects agents based on structural importance, AgentVerse~\cite{agentverse2024} explores emergent behavioral patterns in groups, and MacNet~\cite{macnet2025} and CARD~\cite{card2026} pattern agent interactions after specialized multi-agent structures and conditional topologies. Despite their remarkable designs, existing methods largely overlook group memory mechanisms: they discard interaction histories after each task and fail to preserve or leverage shared strategic knowledge. In this work, we focus on developing group memory for MAS to enable continual learning and collaboration across tasks.

\begin{figure*}[t]
\centering
\includegraphics[width=\textwidth]{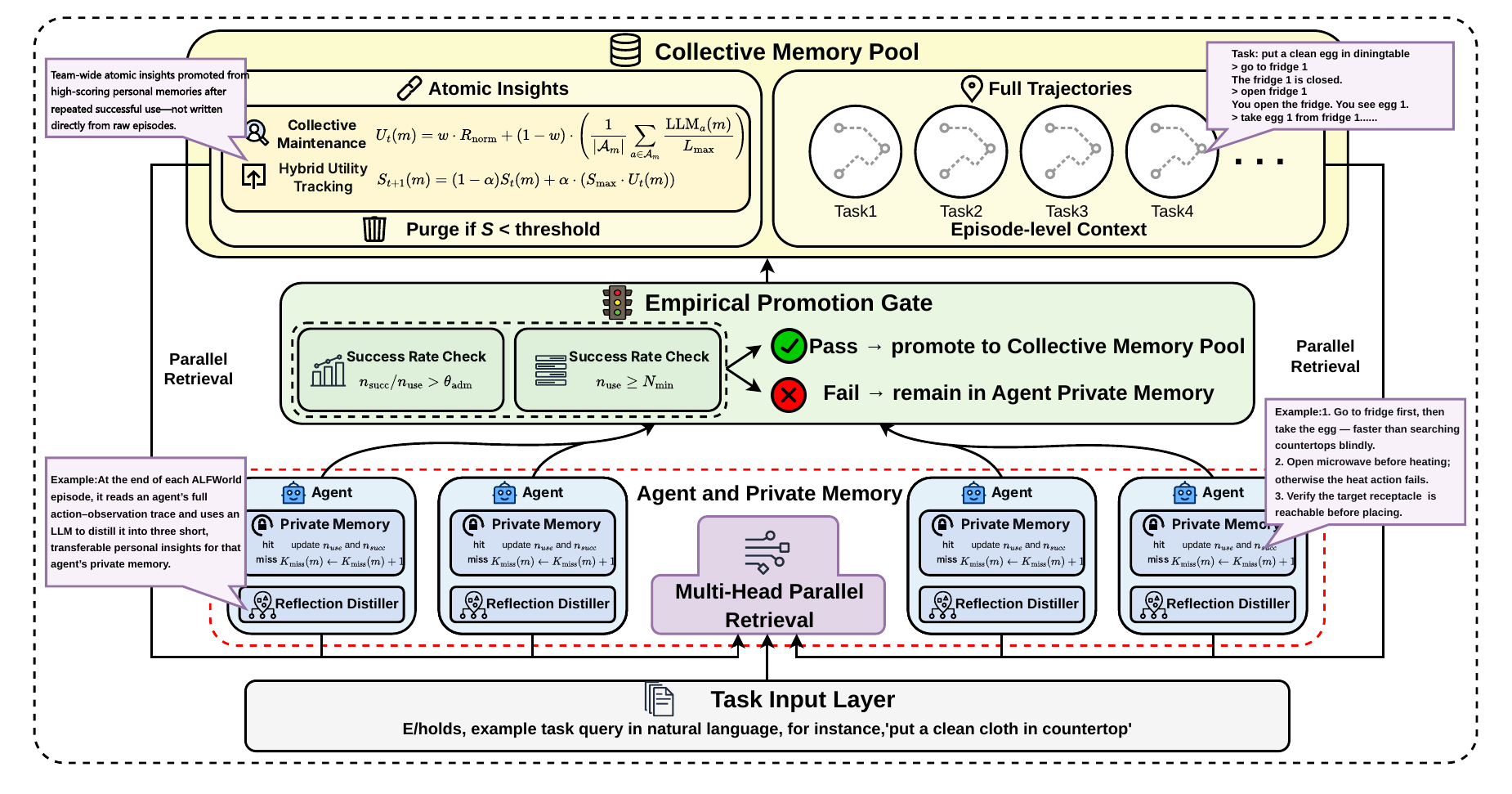}
\vspace{-0.25in}
\caption{The overall structural overview of the CoMem architecture. Agents maintain isolated Private Memory slots and share a Collective Memory Pool via a stringent empirical promotion gate.}
\vspace{-0.2in}
\label{fig:architecture_wide}
\end{figure*}

\subsection{Memory Mechanisms for Autonomous Agents}
Memory serves as a critical pillar for lifelong learning and continuous adaptation in autonomous agents~\cite{cognitive2024, amem2025, mem02025, awm2025, expel2024}. In single-agent setups, frameworks like MemGPT~\cite{memgpt2023} emulate OS-level virtual memory hierarchies to bypass context limitations. Reflexion~\cite{reflexion2023} utilizes linguistic reinforcement to remember structural failures, HippoRAG~\cite{hippo2024} performs neurobiologically inspired associative retrieval, and Voyager~\cite{voyager2023} develops a persistent skill library to navigate open-world environments.  However, translating single-agent memory into MAS poses immense friction due to cross-agent interference. Unrestricted shared configurations, as seen in MemoryBank~\cite{memorybank2024} or simple generative agent shared logs~\cite{generative2023}, lead to massive information dilution. While G-Memory~\cite{gmemory2025} attempts to structure this data using multi-layered graphs, it maintains a flat read/write structure across agents. Recent works have highlighted the critical drawbacks of such shared architectures, such as information contamination~\cite{wang2026, bras2026} and context homogenization~\cite{zhang2025}. This paper overcomes this limitation by upgrading the memory mechanism to group memory~\cite{rezazadeh2025, du2026}.

% %%%%%%%%%%%%%%%%%%%%%%%%%%%%%%%%%%%%%%%%%%%%%%%%%%%%%
% SECTION 3: METHODOLOGY
% %%%%%%%%%%%%%%%%%%%%%%%%%%%%%%%%%%%%%%%%%%%%%%%%%%%%%
\section{Methodology}
\subsection{Overview}

Let $\mathcal{T} = \{T_1, T_2, \dots, T_N\}$ be a sequence of collaborative tasks drawn from a dynamic execution environment. A MAS consisting of a set of autonomous agents $\mathcal{A} = \{a_1, a_2, \dots, a_{|\mathcal{A}|}\}$ is deployed to solve these tasks sequentially. For each task $T_k$, the agents collaborate over an episodic horizon of execution steps.

During the execution of task $T_k$, an agent $a_i \in \mathcal{A}$ generates a localized execution trace
\[
\tau_i^k = \{(s_1, x_1, y_1), \dots, (s_t, x_t, y_t)\},
\]
which encapsulates environmental states $s$, individual chain-of-thought rationales $x$, and executed actions $y$. 

In conventional shared-memory MAS frameworks, all agents indiscriminately write their episodic summaries into a monolithic, flat collective repository $\mathcal{M}_{\text{shared}}$. We formalize the \textit{Memory Pollution Problem} as follows~\cite{wang2026, bras2026}: Let $\mathcal{M}_t$ be the state of the shared memory at time step $t$. The quality of $\mathcal{M}_t$ degrades over time due to two primary noise sources:

(1) \textbf{Execution Noise Dilution}: unverified trial-and-error behaviors and circumstantial successes contaminate the repository~\cite{wang2026}.

(2) \textbf{Specialization Homogenization}: cross-agent overwriting forces role-specific cognitive trajectories to conform to a noisy average~\cite{zhang2025}, such that
\[
\lim_{t \to \infty} H(\pi_{a_i} \parallel \pi_{a_j}) = 0, \quad \forall a_i, a_j \in \mathcal{A}.
\]

The objective of CoMem is to maintain a decoupled memory space
\[
\mathcal{M} = \left\langle \{\mathcal{M}_{\text{priv}}^{(i)}\}_{i=1}^{|\mathcal{A}|}, \mathcal{M}_{\text{coll}} \right\rangle
\]
such that individual learning purity is preserved in the isolated private stores $\mathcal{M}_{\text{priv}}^{(i)}$~\cite{rezazadeh2025}, while organization-level general wisdom is filtered and sustained in the pristine global store $\mathcal{M}_{\text{coll}}$~\cite{du2026}.

Figure~\ref{fig:architecture_wide} illustrates the overall structure of CoMem. At its core, CoMem maintains two logically and physically isolated memory tiers: a Private Layer and a Collective Pool. The Private Layer consists of exclusive collections, each dedicated to a single agent, ensuring that no cross-agent contamination can occur at the storage level. The Collective Pool is a shared repository that stores only those insights that have passed a rigorous empirical validation gateway.

The lifecycle of each memory unit is governed by a unified metadata schema. Every memory entry $m$ is stored as a structured tuple:
\[
m = \langle \text{doc\_content}, \text{metadata} \rangle
\]
where $\text{metadata} = \{ \text{agent\_name}, S, n_{\text{use}}, n_{\text{succ}}, \iota_{\text{prom}} \}$, representing the owner agent, utility score, invocation counter, successful invocation counter, and promotion status, respectively.

Memory flows through the system along a bidirectional pathway. New experiences are first generated by individual agents and stored in their private collections. These entries undergo rolling pruning based on usage recency: frequently retrieved entries are retained, while entries that remain unused beyond a threshold are systematically evicted. When an entry demonstrates sufficient empirical success, it qualifies for promotion and is replicated into the Collective Pool. Promoted entries are then accessible to all agents, while collective entries that fail to maintain their utility over time are purged from the pool. This design ensures that the Private Layer remains compact and role-specific, while the Collective Pool evolves into a curated repository of generalizable, high-value organizational wisdom. 

Next, we introduce the core components  Private Experience Sedimentation, Collective Wisdom Curation and Parallel Dual-Stream Retrieval in detail.

\subsection{Private Experience Sedimentation}

Building on the layered memory architecture outlined above, we first focus on how each agent preserves and refines its personal learning trajectory. At the conclusion of every episode, agent $a_i$ extracts its own execution trace $\tau_i^k$ from the sequence of runtime states and decisions. Rather than storing these raw, noisy trajectories directly, $a_i$ engages in a structured reflection process\cite{reflexion2023, expel2024}: a dedicated LLM-based distiller, operating from the agent’s unique perspective, transforms the episodic trace into a small set of succinct, actionable textual insights. Each insight distills a strategic lesson or behavioral pattern that captures individual learning value.

These distilled insights are then stored in the agent’s private memory collection, $\mathcal{M}_{\text{priv}}^{(i)}$, each assigned an initial utility score $S_0$. Associated metadata counters—including usage count $n_{\text{use}}$, success count $n_{\text{succ}}$, and a promotion flag $\iota_{\text{prom}}$—are initialized to their default states, indicating that the entry is newly created and yet to be validated or considered for promotion. Importantly, $\mathcal{M}_{\text{priv}}^{(i)}$ enforces strict access control by filtering entries based on agent identity; this structural separation guarantees that no cross-agent contamination can occur, and preserves the integrity of each agent’s own reflective history.

To maintain the efficiency and relevance of private memory, each agent employs a rolling pruning strategy for non-promoted entries\cite{memorybank2024, mem02025}. Specifically, a consecutive miss counter $K_{\text{miss}}(m)$ is assigned to every private memory item $m$. After each episode, entries are evaluated for activity: if an insight $m$ was retrieved and injected into a prompt during decision-making, $K_{\text{miss}}(m)$ is reset to zero; otherwise, it is incremented by one. Once an entry’s miss counter surpasses a predefined threshold $K_{\text{max}}$, it is permanently removed from the private collection. This lightweight mechanism allows private memories to remain concise and focused by naturally filtering out stale or unused knowledge, even in the sparse and uncertain feedback setting typical of multi-agent environments.

Through this private sedimentation process, CoMem ensures that each agent can robustly preserve and refine its unique experiential base, laying the groundwork for both long-term individual expertise and reliable contributions to collective memory.

\subsection{Collective Wisdom Curation}

Building upon the foundation of private experience sedimentation, CoMem ensures that only truly robust and generalizable insights are elevated to the organizational memory. To this end, we introduce a carefully designed empirical promotion gateway that rigorously filters candidate knowledge before admitting it into the collective pool.

For each private memory entry $m$ generated by agent $a_i$, promotion is governed by an empirical, baseline-conditioned rule:
\[
\iota_{\text{prom}} = \mathbb{I}\Big[ \big(N_{\text{use}}(m) \ge N_{\text{min}}\big) \,\land\, \big(\bar{R}_m \ge \bar{R}\big) \Big]
\]
where $N_{\text{use}}(m)$ counts the total number of episodes in which $m$ has been successfully retrieved—enforcing a statistical floor $N_{\text{min}}$ to guard against incidental or anomalous activation. The empirical value $\bar{R}_m$ captures the average task score achieved in episodes where $m$ was actively used; $\bar{R}$ serves as an environment-wide moving baseline of historical average performance. This conjunctive criterion ensures that only entries demonstrating both sufficient exposure and consistent, above-average impact are promoted. Upon satisfaction, $m$ is replicated to the collective memory $\mathcal{M}_{\text{coll}}$ and removed from the originating private store, avoiding redundancy while underpinning group knowledge with individually vetted experience\cite{rezazadeh2025, gmemory2025}.

Once admitted, collective entries are not static: each is subject to continuous quality control via a hybrid utility assessment. At the end of every episode, for each retrieved collective entry, we compute a comprehensive utility score $U_t(m)$:
\[
U_t(m) = w \cdot R_{\text{norm}} + (1 - w) \cdot \left( \frac{1}{|\mathcal{A}_m|} \sum_{a \in \mathcal{A}_m} \frac{\operatorname{LLM}_a(m)}{L_{\text{max}}} \right),
\]
where $R_{\text{norm}}$ is the normalized episodic reward, $\mathcal{A}_m$ is the subset of agents evaluating $m$, $\operatorname{LLM}_a(m)$ is a subjective score from offline LLM aggregation, and $w$ sets the weighting between objective performance and subjective insight quality. 

The long-term utility $S_{t}(m)$ is updated as an exponential moving average:
\[
S_{t+1}(m) = (1 - \alpha) S_t(m) + \alpha \cdot \left( S_{\text{max}} \cdot U_t(m) \right),
\]
with smooth score adjustment via EMA update rate $\alpha$ and capped at $S_{\text{max}}$. Any entry whose score persistently drops below an adaptive threshold is systematically removed, ensuring that the collective memory remains lean, relevant, and filled only with organizational wisdom proven valuable over time.

Through this stringent curation process, CoMem’s collective pool matures into a trusted knowledge base, continuously distilled and refined by empirical feedback—ensuring that only generalizable and enduring insights guide future multi-agent collaboration.

\begin{table*}[t]
\centering
\footnotesize
\setlength{\tabcolsep}{3pt}
\vspace{-0.1in}
\label{tab:1}
\begin{tabular}{lcc|cc|cc|cc}
\toprule
\multirow{2}{*}{\textbf{Memory Mechanism}}
  & \multicolumn{2}{c|}{\textbf{AutoGen}}
  & \multicolumn{2}{c|}{\textbf{DyLAN}}
  & \multicolumn{2}{c|}{\textbf{MacNet}}
  & \multicolumn{2}{c}{\textbf{CARD}} \\
\cmidrule(lr){2-3}\cmidrule(lr){4-5}\cmidrule(lr){6-7}\cmidrule(lr){8-9}
& \textbf{ALFWorld} & \textbf{PDDL}
& \textbf{ALFWorld} & \textbf{PDDL}
& \textbf{ALFWorld} & \textbf{PDDL}
& \textbf{ALFWorld} & \textbf{PDDL} \\
\midrule
No-memory
  & 79.10\textsuperscript{\tiny +0.00} & 69.52\textsuperscript{\tiny +0.00}
  & 80.60\textsuperscript{\tiny +0.00} & 67.78\textsuperscript{\tiny +0.00}
  & 79.85\textsuperscript{\tiny +0.00} & 60.78\textsuperscript{\tiny +0.00}
  & 84.33\textsuperscript{\tiny +0.00} & 70.01\textsuperscript{\tiny +0.00} \\
Voyager
  & 76.87\textsuperscript{\tiny -2.23} & 67.60\textsuperscript{\tiny -1.92}
  & 72.39\textsuperscript{\tiny -8.21} & 58.17\textsuperscript{\tiny -9.61}
  & 77.61\textsuperscript{\tiny -2.24} & 52.07\textsuperscript{\tiny -8.71}
  & 87.31\textsuperscript{\tiny +2.98} & 51.92\textsuperscript{\tiny -18.09} \\
MemoryBank
  & 70.90\textsuperscript{\tiny -8.20} & 53.45\textsuperscript{\tiny -16.07}
  & 71.64\textsuperscript{\tiny -8.96} & 50.92\textsuperscript{\tiny -16.86}
  & 73.13\textsuperscript{\tiny -6.72} & 50.94\textsuperscript{\tiny -9.84}
  & 77.61\textsuperscript{\tiny -6.72} & 49.89\textsuperscript{\tiny -20.12} \\
Generative
  & 72.39\textsuperscript{\tiny -6.71} & 61.29\textsuperscript{\tiny -8.23}
  & 65.67\textsuperscript{\tiny -14.93} & 70.34\textsuperscript{\tiny +2.56}
  & 73.13\textsuperscript{\tiny -6.72} & 59.96\textsuperscript{\tiny -0.82}
  & 90.30\textsuperscript{\tiny +5.97} & 58.00\textsuperscript{\tiny -12.01} \\
MetaGPT
  & 65.67\textsuperscript{\tiny -13.43} & 69.85\textsuperscript{\tiny +0.33}
  & 67.16\textsuperscript{\tiny -13.44} & 52.80\textsuperscript{\tiny -14.98}
  & 71.64\textsuperscript{\tiny -8.21} & 50.83\textsuperscript{\tiny -9.95}
  & 90.30\textsuperscript{\tiny +5.97} & 53.53\textsuperscript{\tiny -16.48} \\
ChatDev
  & 75.37\textsuperscript{\tiny -3.73} & 61.58\textsuperscript{\tiny -7.94}
  & 79.85\textsuperscript{\tiny -0.75} & 68.66\textsuperscript{\tiny +0.88}
  & 76.12\textsuperscript{\tiny -3.73} & 61.95\textsuperscript{\tiny +1.17}
  & 76.12\textsuperscript{\tiny -8.21} & 66.52\textsuperscript{\tiny -3.49} \\
G-Memory
  & 72.39\textsuperscript{\tiny -6.71} & 72.05\textsuperscript{\tiny +2.53}
  & 71.64\textsuperscript{\tiny -8.96} & 59.53\textsuperscript{\tiny -8.25}
  & 84.33\textsuperscript{\tiny +4.48} & 57.08\textsuperscript{\tiny -3.70}
  & 91.04\textsuperscript{\tiny +6.71} & 56.62\textsuperscript{\tiny -13.39} \\
\midrule
\textbf{CoMem (Ours)}
  & \textbf{88.81}\textsuperscript{\tiny +9.71} & \textbf{74.44}\textsuperscript{\tiny +4.92}
  & \textbf{87.31}\textsuperscript{\tiny +6.71} & \textbf{69.50}\textsuperscript{\tiny +1.72}
  & \textbf{89.55}\textsuperscript{\tiny +9.70} & \textbf{70.19}\textsuperscript{\tiny +9.41}
  & \textbf{90.30}\textsuperscript{\tiny +5.97} & \textbf{72.74}\textsuperscript{\tiny +2.73} \\
\bottomrule
\end{tabular}
\caption{Success rates on ALFWorld and PDDL across different MAS frameworks. Superscripts indicate absolute percentage-point differences from the No-memory baseline. The underlying LLM backbone is DeepSeek-V4-Flash.}
\end{table*}

\subsection{Parallel Dual-Stream Retrieval}

Building on the stratified structure of both private and collective memories, CoMem empowers agents to leverage the strengths of both perspectives during task execution. When a new task initialization query arrives, each agent simultaneously retrieves information from the two isolated repositories through a parallel dual-stream mechanism. This design not only eliminates the bottleneck of sequential lookups, but also ensures that the agent’s decision context is enriched by both individualized tactical know-how and distilled organizational wisdom—allowing for synergy without cross-tier contamination.

\paragraph{Private Stream Retrieval.}
The private retrieval stream zeroes in on the agent’s own archive of experience. Using semantic embedding similarity\cite{lewis2020rag}, the top-ranked private memory entries—those most relevant to the current task context—are rapidly retrieved from the agent's individual tier. Injecting these insights directly into the agent's prompt equips the agent with tailored, personal behavioral guidance honed across its own learning history, supporting both specialization and agility in complex environments.

\paragraph{Collective Stream Retrieval and Diversity Filtering.}
Concurrently, the collective stream taps into the shared pool of group-sourced wisdom, retrieving entries that have been empirically validated and promoted from across the entire team. To address the risk of semantic overlap—inevitable when multiple agents converge on similar insights—CoMem integrates an online hierarchical clustering filter\cite{murtagh2012}. By grouping candidates based on semantic distance and selecting the most representative entry from each cluster, the mechanism constructs a consensus context that is both succinct and diverse. This not only minimizes redundant knowledge within the prompt, conserving valuable token space, but also ensures that agents benefit from a structurally rich set of perspectives distilled from the broader collective experience.

Through parallel dual-stream retrieval, CoMem instantiates a memory interface where individual expertise and shared knowledge dynamically inform each agent’s reasoning process. This architecture tightly couples personal exploration with the evolving consensus of the group, fostering both innovation and reliability in multi-agent coordination.

% %%%%%%%%%%%%%%%%%%%%%%%%%%%%%%%%%%%%%%%%%%%%%%%%%%%%%
% SECTION 4: EXPERIMENTS
% %%%%%%%%%%%%%%%%%%%%%%%%%%%%%%%%%%%%%%%%%%%%%%%%%%%%%

\section{Experiments}

In this section, we conduct experiments to answer:
\textbf{(RQ1)} How does CoMem perform compared to single-/multi-agent memory architectures?
\textbf{(RQ2)} Does CoMem incur excessive resource overhead?
\textbf{(RQ3)} How sensitive is CoMem to its key components and hyper-parameters?

\subsection{Experiment Setup}

\paragraph{Datasets and Baselines.}
We evaluate CoMem on two standard decision-making benchmarks:
ALFWorld~\cite{alfworld2021} and PDDL~\cite{pddl1998}.
ALFWorld tests embodied multi-step control with dense environmental feedback,
while PDDL requires long-horizon symbolic planning and strict action syntax. We compare against No-memory, four single-agent memory methods adapted to MAS
(Voyager~\cite{voyager2023}, MemoryBank~\cite{memorybank2024},
Generative Agents~\cite{generative2023}),
and three cooperative shared-memory schemes
(MetaGPT~\cite{metagpt2024}, ChatDev~\cite{chatdev2024}, G-Memory~\cite{gmemory2025}).
All memory modules are plugged into the host MAS without modifying its native coordination protocol.

\paragraph{MAS Frameworks and LLM Backbones.}
Experiments are conducted on four representative MAS frameworks:
AutoGen~\cite{autogen2024}, DyLAN~\cite{dylan2024},
MacNet~\cite{macnet2025}, and CARD~\cite{card2026}.
We adopt DeepSeek-V4-Flash~\cite{deepseek2026} as the primary backbone,
with Qwen3-32B~\cite{qwen2025} for supplementary validation. DeepSeek-V4-Flash represents a large-scale LLM backbone with a sparse architecture, while Qwen3-32B serves as a dense transformer backbone for supplementary validation. In our framework, the backbone LLM is only responsible for general reasoning and generation, whereas memory organization, evolution, and retrieval are explicitly managed by CoMem. Therefore, the effectiveness of CoMem does not depend on any specific internal architecture of the underlying LLM and can be integrated with diverse backbone models. To verify this property, we additionally evaluate CoMem with the dense Qwen3-32B backbone and observe consistent improvements, demonstrating that CoMem is backbone-agnostic and generalizes across different model families.

\paragraph{Parameter Configurations.}
Unless otherwise stated, CoMem uses retrieval depth $k{=}3$ for both personal and collective pools,
private rolling window size $K_{\text{max}}{=}5$ (PDDL) / $6$ (ALFWorld),
collective forgetting threshold $30$,
collective score EMA $\alpha{=}0.5$, and reward--utility mixing weight $w{=}0.5$.
Promotion requires a minimum usage count of $3$ and above-baseline mean reward ($\bar{R}_m \ge \bar{R}$).
Detailed ablations are reported in Section~\ref{sec:framework-analysis}.

\begin{figure}[t]
    \centering
    \includegraphics[width=1\linewidth]
    {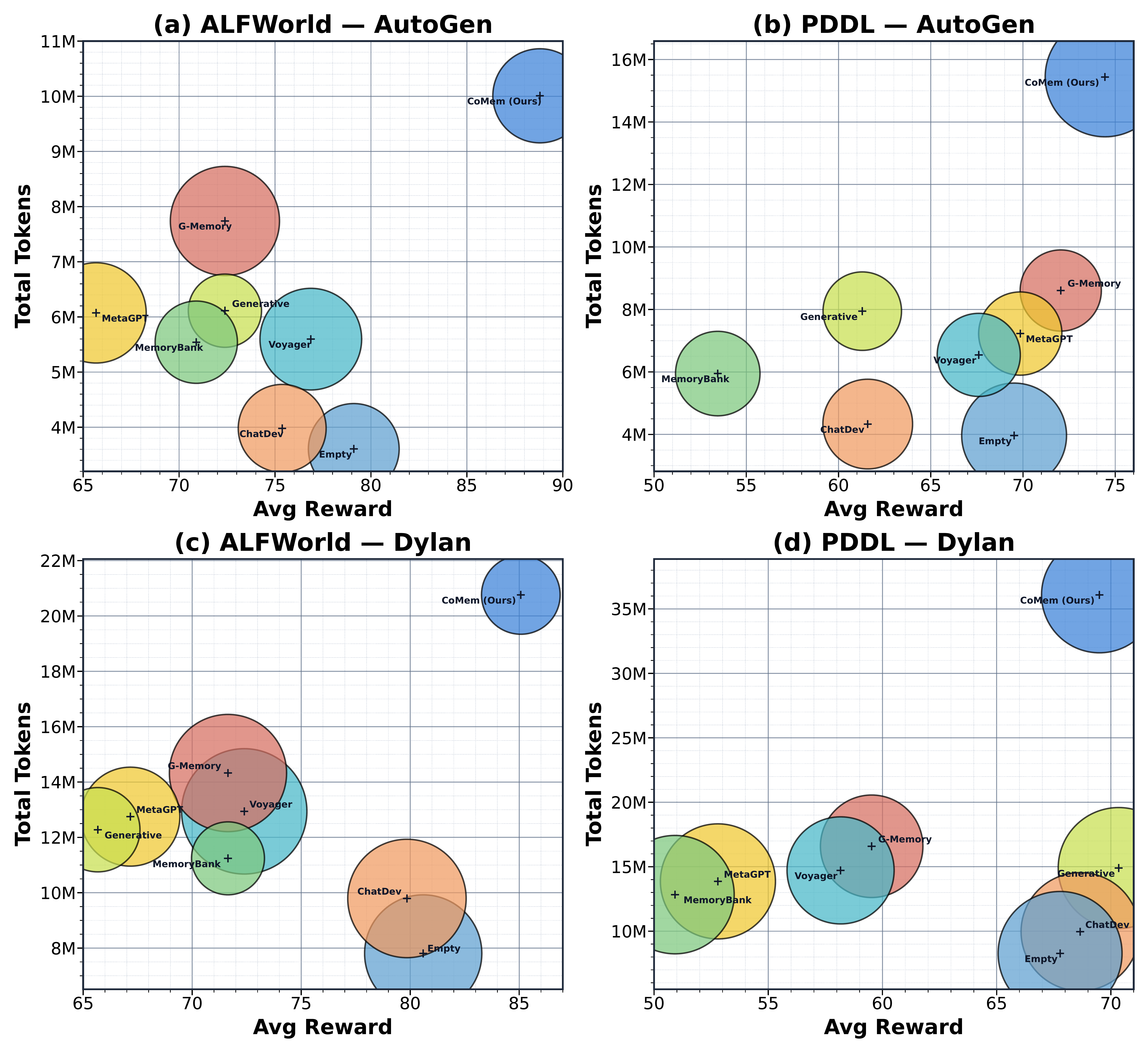}
    \vspace{-0.2in}
    \caption{Performance versus token cost across memory mechanisms.}
    \label{fig:cost_analysis}
    \vspace{-0.2in}
\end{figure}

% =====================================================
\subsection{Main Results (RQ1)}

Table 1 reports comprehensive success rates across all configurations.
We summarize the key observations as follows.

\paragraph{Takeaway 1:}
CoMem consistently achieves strong performance across frameworks.
On MacNet, CoMem elevates success rates from 79.85 to 89.55 on ALFWorld
and from 60.78 to 70.19 on PDDL.
Consistent improvements are observed across AutoGen, DyLAN, and CARD,
with average gains of 8.02 on ALFWorld and 4.70 on PDDL,
computed as the mean of absolute percentage-point increases over the No-memory baseline.
Notably, even on the strong CARD backbone, where the No-memory baseline already achieves 84.33 on ALFWorld and 70.01 on PDDL,
CoMem still delivers non-trivial improvements of 90.30 and 72.74,
further validating its effectiveness as a plug-and-play memory layer that consistently enhances performance regardless of the underlying MAS strength.

\paragraph{Takeaway 2:}
Conventional cooperative shared memory frequently underperforms the No-memory baseline.
For instance, under DyLAN on PDDL, G-Memory yields 59.53, significantly below No-memory's 67.78;
MetaGPT drops from 67.78 to 52.80.
Single-agent memory methods adapted to MAS, such as Voyager and MemoryBank, also exhibit inconsistent behavior,
improving marginally in some configurations while degrading in others.
This counterintuitive pattern arises from indiscriminate cross-agent read/write, causing noise accumulation and behavioral homogenization.
In a shared repository without structural isolation, a large volume of unverified trial-and-error records is continuously written,
progressively degrading storage quality---a phenomenon we refer to as noise accumulation.
Simultaneously, strategic trajectories of different roles are repeatedly overwritten and averaged out during cross-agent interactions,
gradually diminishing the value of role differentiation---a phenomenon we refer to as behavioral homogenization.
These two degradation patterns are consistent with the memory pollution problem formalized in Section 3.
CoMem prevents individual experiences from being contaminated through structured isolation of the private layer,
and ensures that only sufficiently validated insights can enter the collective pool via the empirical promotion gateway,
thereby effectively suppressing both degradation pathways.

\subsection{Cost Analysis (RQ2)}

Figure~\ref{fig:cost_analysis} visualizes the performance--token trade-off.

\paragraph{Takeaway 3:}
CoMem achieves the highest success rates with a modest increase in token consumption,
primarily attributable to parallel dual-stream retrieval and empirical promotion validation.
This overhead translates into substantial and consistent performance gains,
with average absolute improvements of 8.02 on ALFWorld and 4.70 on PDDL.
In contrast, cooperative baselines often incur comparable or even higher token costs without yielding consistent performance benefits,
further underscoring CoMem's favorable cost-effectiveness profile.

% =====================================================
\subsection{Framework Analysis (RQ3)}
\label{sec:framework-analysis}
% =====================================================

\begin{figure}[t]
    \centering
    \includegraphics[width=0.95\linewidth]{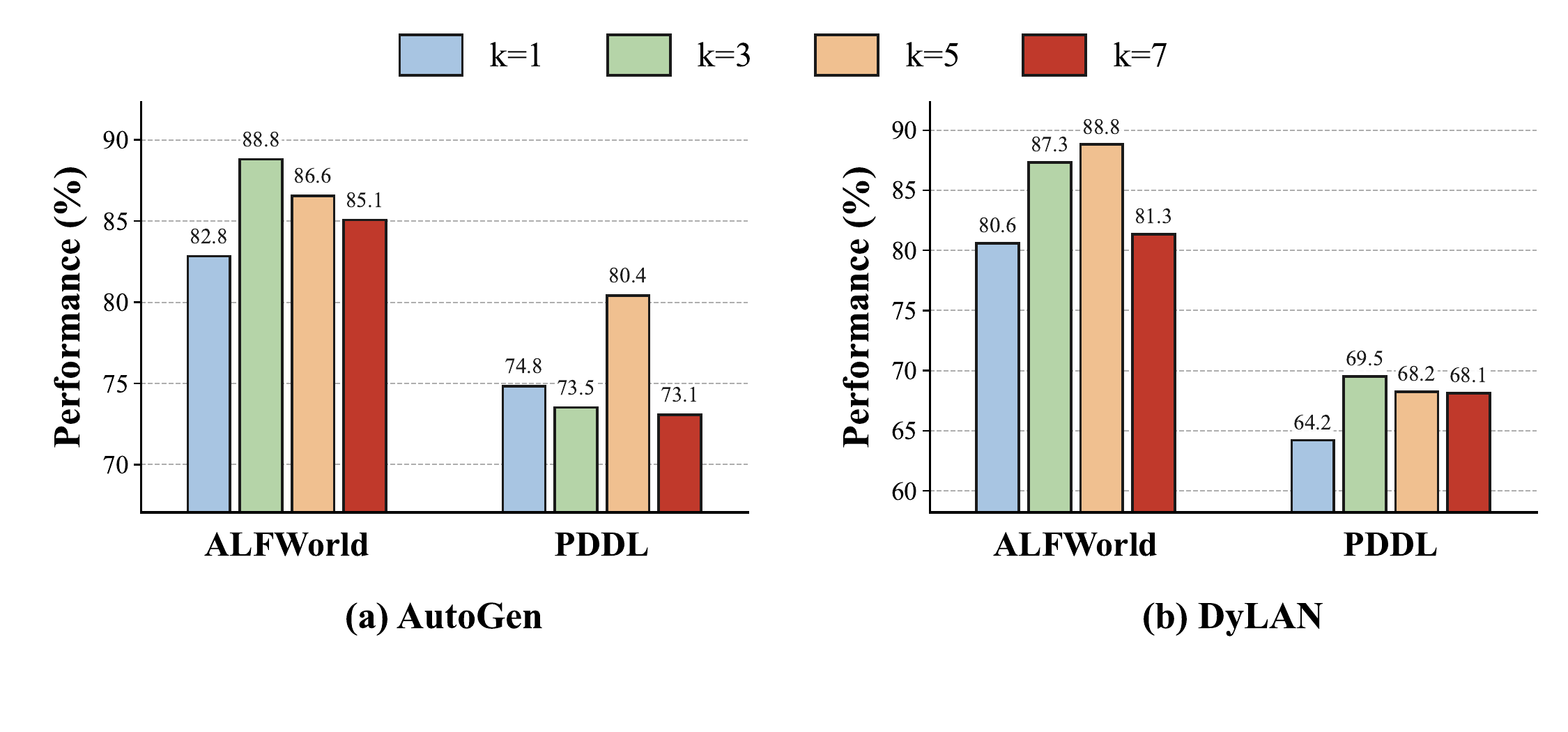}
   \vspace{-0.2in}
    \caption{Impact of retrieval depth $k$ on system performance.}
    \label{fig:ablation_topk}
    \vspace{-0.2in}
\end{figure}

We systematically evaluate CoMem's retrieval depth, component necessity, and lifecycle hyper-parameters. All ablation trials are baseline-controlled and executed with the AutoGen and DyLAN frameworks on both benchmarks.

\paragraph{Retrieval Depth Sensitivity.}
We ablate the operational retrieval depth $k \in \{1, 3, 5, 7\}$ for both personal and collective pools. Figure~\ref{fig:ablation_topk} shows domain-dependent optima:
on AutoGen, ALFWorld peaks at $k{=}3$ (88.80) while PDDL peaks at $k{=}5$ (80.40);
on DyLAN, ALFWorld peaks at $k{=}5$ (88.80) and PDDL at $k{=}3$ (69.50).
Moderate $k \in \{3,5\}$ yields robust performance; larger $k$ injects noise, smaller $k$ misses context.
Notably, the gap between $k{=}1$ (74.80) and $k{=}5$ (80.40) is substantial (5.60 points),
demonstrating that retrieval scaling is contingent upon task difficulty.
These results suggest that optimal retrieval depth varies across both backbone frameworks and task domains,
and that a moderate retrieval budget ($k \in \{3, 5\}$) consistently yields robust performance.

\paragraph{Memory Granularity and Component Necessity.}
To confirm whether the two-tiered paradigm is structurally necessary, we isolate the Personal Layer and Collective Pool on both benchmarks.
In the w/o Collective configuration, cross-task empirical promotion is disabled (personal-only);
in the w/o Personal configuration, trajectories are distilled directly into the shared pool at episode termination without individual buffering (collective-only).

\begin{table}[t]
  \centering
  \label{tab:component-ablation}
  \setlength{\tabcolsep}{6pt}
  \renewcommand{\arraystretch}{1.15}
  \begin{tabular}{l c c c c}
    \toprule
    MAS & Private & Collective & ALFWorld & PDDL \\
    \midrule
    \multirow{3}{*}{AutoGen}
      & \cmark &  & 83.58 & 62.41 \\
      &  & \cmark & 85.07 & 65.10 \\
      & \cmark & \cmark & \textbf{88.31} & \textbf{74.44} \\
    \midrule
    \multirow{3}{*}{DyLAN}
      & \cmark & & 80.25 & 59.17 \\
      &  & \cmark & 77.61 & 67.22 \\
      & \cmark & \cmark & \textbf{87.31} & \textbf{69.50} \\
    \bottomrule
  \end{tabular}
  \caption{Component ablation of CoMem on PDDL and ALFWorld.}
\end{table}

Table 2 reveals that the full CoMem setup consistently outperforms either isolated variant.
On AutoGen with PDDL, the full configuration achieves 74.44, surpassing collective-only (65.10) and personal-only (62.41).
The performance gap between personal-only and collective-only suggests that maintaining individual episodic buffers is more fundamental to preserving operational integrity than immediately pooling unvalidated collaborative knowledge.
Critically, removing the collective layer leads to a 12.03-point drop on AutoGen with PDDL (74.44 $\rightarrow$ 62.41),
while removing the private layer causes a 9.34-point drop (74.44 $\rightarrow$ 65.10).
This asymmetry indicates that both layers contribute essential yet distinct functions,
with the private layer providing a slightly stronger safeguard against memory pollution.
\paragraph{Threshold Sensitivity: Private Rolling Window and Collective EMA.}
We conduct a fine-grained sweep on AutoGen across two key dimensions:
(i)~the private rolling window size $K_{\text{max}}$, which governs the eviction of stale private entries upon consecutive non-activation,
and (ii)~the collective EMA update rate $\alpha$, which controls how rapidly collective memory scores adapt to newly observed performance.
Results are in Table~\ref{tab:sensitivity}.

\begin{table}[htbp]
\centering
\small 
\setlength{\tabcolsep}{6pt}  
\begin{tabular}{c c c | c c c}
\toprule
\multicolumn{3}{c}{\textbf{Pruning ($K_{\max}$)}} & \multicolumn{3}{c}{\textbf{Fusion ($\alpha$)}} \\
\cmidrule(lr){1-3} \cmidrule(lr){4-6}
$K_{\max}$ & ALFWorld & PDDL & $\alpha$ & ALFWorld & PDDL \\
\midrule
3 & 84.33 & 66.21 & 0.3 & 81.34 & 67.80 \\
4 & 82.09 & 70.80 & 0.4 & 85.07 & 71.06 \\
5 & 84.33 & \textbf{74.44} & 0.5 & \textbf{88.81} & \textbf{74.44} \\
6 & \textbf{88.81} & 71.14 & 0.6 & 85.82 & 72.41 \\
7 & 87.31 & 71.57 & 0.7 & 80.60 & 69.53 \\
8 & 85.07 & 67.33 & 0.8 & 84.33 & 65.89 \\
\bottomrule
\end{tabular}
\caption{Sensitivity analysis of private rolling window size $K_{\text{max}}$ and collective EMA update rate $\alpha$ on PDDL and ALFWorld success rates.}
\label{tab:sensitivity}
\end{table}

\paragraph{Private Rolling Window Size ($K_{\text{max}}$).}
We ablate $K_{\text{max}} \in \{3, 4, 5, 6, 7, 8\}$ on AutoGen and observe an inverted-U relationship on both benchmarks.
For PDDL, the optimal window size is $K_{\text{max}}=5$, yielding the best success rate of 74.44.
When the window is too small ($K_{\text{max}}=3$), performance drops to 66.21 due to over-pruning of infrequently retrieved but potentially useful experiences.
When the window is excessively large ($K_{\text{max}}=8$), performance declines to 67.33 as task-specific noise accumulates and degrades the signal-to-noise ratio.
For ALFWorld, the optimal window size shifts to $K_{\text{max}}=6$, achieving 88.81.
ALFWorld exhibits a flatter performance curve than PDDL, suggesting less sensitivity to private memory pruning than the structurally complex planning domain.
The suboptimal performance at $K_{\text{max}}=4$ (82.09) and $K_{\text{max}}=7$ (87.31) further confirms that moderate window size strikes the best trade-off between retaining useful experiences and eliminating obsolete knowledge.

\paragraph{Collective EMA Update Rate ($\alpha$).}
We ablate $\alpha \in \{0.3, 0.4, 0.5, 0.6, 0.7, 0.8\}$ on AutoGen and observe a consistent inverted-U trend on both benchmarks.
The momentum coefficient $\alpha$ governs the collective score update rule
$S_{t+1} = (1-\alpha)S_t + \alpha \cdot (100 \cdot U_t)$,
where a smaller $\alpha$ emphasizes historical scores while a larger $\alpha$ favors newly observed performance.
When $\alpha$ is too small ($\alpha=0.3$), collective scores lag behind environmental changes, yielding 67.80 on PDDL and 81.34 on ALFWorld, failing to promptly deprecate entries whose utility has diminished.
When $\alpha$ is too large ($\alpha=0.8$), the system overreacts to single-trial anomalies, yielding 65.89 on PDDL and 84.33 on ALFWorld, promoting transient noise into the collective pool.
The optimal equilibrium is achieved at $\alpha=0.5$ on both benchmarks (PDDL: 74.44; ALFWorld: 88.81), where the system maintains sufficient responsiveness while effectively suppressing noise.
Across all tested $\alpha$ values, ALFWorld consistently outperforms PDDL, and both benchmarks exhibit the same optimal EMA setting, confirming that $\alpha=0.5$ is a robust default across task domains.

% %%%%%%%%%%%%%%%%%%%%%%%%%%%%%%%%%%%%%%%%%%%%%%%%%%%%%
% SECTION 6: CONCLUSION
% %%%%%%%%%%%%%%%%%%%%%%%%%%%%%%%%%%%%%%%%%%%%%%%%%%%%%

\section{Conclusion}

This paper introduces CoMem, a group memory architecture designed to overcome the memory pollution problem in LLM-based multi-agent systems. By structurally separating agents' private experiences from the collective memory pool and implementing a rigorous empirical promotion mechanism, CoMem preserves individual learning while ensuring that only validated, high-value insights are shared at the organizational level. A parallel dual-stream retrieval strategy allows each agent to leverage both personal knowledge and generalized collective wisdom for improved collaboration. Comprehensive experiments demonstrate that CoMem provides a robust and effective solution for mitigating memory pollution in LLM-based multi-agent systems.

\fontsize{9.5pt}{10.5pt} \selectfont
\bibliography{references}

\end{document}